\documentclass[runningheads]{llncs}
\usepackage{amsmath}
\usepackage[T1]{fontenc}
\usepackage{graphicx,verbatim}
\graphicspath{{graphics/}}
\usepackage{hyperref}
\usepackage{algorithm}
\usepackage{algorithmic}
\usepackage{amssymb}
\usepackage{booktabs}
\usepackage{placeins}
\usepackage{orcidlink}
\renewcommand{\orcidID}[1]{\,\orcidlink{#1}}

\usepackage{color}

\newcommand{\iv}{IV}
\newcommand{\ivs}{IVs}

\begin{document}
\title{ML-CLAPPS: A Probabilistic Framework for Imbalanced Multilabel Learning under Class Prioritization}
\titlerunning{ML-CLAPPS: Imbalanced Multilabel Learning with Class Prioritization}

\author{Priel Hazan\orcidID{0009-0000-0688-8387} \and
Nir Almog\orcidID{0009-0001-7484-7912} \and
Noam Keidar\orcidID{0000-0002-6848-6428} \and
Yael Yaniv\orcidID{0000-0002-5183-6284} \and
Assaf Schuster\orcidID{0000-0002-3311-6937}}
\authorrunning{P. Hazan et al.}
\institute{Technion -- Israel Institute of Technology, Haifa, Israel\\
\email{priel.hazan@campus.technion.ac.il}}
  
\maketitle%
\begin{abstract}

Real-world datasets often exhibit severe class imbalance, degrading the performance of deep learning models, particularly on rare classes. While imbalance is well studied in binary and multiclass settings, it is substantially more challenging in multilabel problems due to complex label co-occurrence.
In clinical applications such as ECG classification, this challenge is coupled with a second critical challenge: the need to prioritize high-risk conditions, regardless of their prevalence, to mitigate the high cost of false negatives.
We introduce the Multilabel Class-Prioritized Probabilistic Sampler (ML-CLAPPS), a learnable sampler that optimizes a distribution over training instances to control the model’s exposure to different classes. This approach addresses multilabel imbalance while enabling prioritization of significant conditions via a user-defined importance vector (\iv). Empirical evaluation on a real-world multilabel ECG image dataset demonstrates that ML-CLAPPS achieves state-of-the-art performance. We show that it aligns class exposure with the specified \iv, and that increasing a class’s importance leads to corresponding improvements in its classification performance, enabling explicit prioritization of clinically critical conditions.
Code at: 
\href{https://github.com/PrielHaz/ml-clapps}{https://github.com/PrielHaz/ml-clapps}.


\end{abstract}

\section{Introduction}

Standard 12-lead electrocardiograms (ECGs) capture cardiac electrical activity from 12 perspectives, enabling the detection of diverse cardiac conditions~\cite{12LeadBraunwald}. In practice, multiple conditions may co-occur within a single recording, requiring multilabel classification approaches~\cite{imbalanced_ecg}.
Training models for this task face two primary hurdles. First, extreme class imbalance, where frequent conditions dominate rare ones~\cite{imbalanced_ecg}. Although it is effectively mitigated in binary or multiclass settings via standard oversampling~\cite{multiclass_profoundly_examined,multiclass_oversampling}, it is considerably more challenging in multilabel learning due to label co-occurrence. Second, varying clinical significance necessitates explicit prioritization of conditions during training~\cite{need_for_giving_importance_to_critical_conditions}.
Standard training pipelines generally treat all instances as equally informative, sampling each with the same frequency. Under severe class imbalance, models tend to neglect rare classes and succumb to the accuracy paradox~\cite{accuracy_paradox}, achieving deceptively high accuracy by consistently predicting the absence of rare classes. Such behavior is unacceptable in clinical settings, as failing to detect a rare but significant condition can have serious consequences~\cite{early_risk_detection}.
To address the imbalance, prior works have employed discrete resampling strategies, such as oversampling minority classes via instance duplication~\cite{LP_ROS,CompareAll,more_duplication_strategies,imbalanced_boosting_and_data_gen,oversampling_most_dominant_imbalance_mitigation_and_imbalance_in_cnns_oversampling,Ando2017DeepOF} and undersampling majority classes~\cite{undersampling}. LP ROS~\cite{LP_ROS,CompareAll} oversamples rare labelsets, ClusterClip~\cite{ClusterClip} samples uniformly across instance clusters with clipping to limit excessive reuse, and LeastSampledClassSampler~\cite{issamemari2021sampler} greedily selects instances from the currently least-exposed class. However, these heuristic-based and discrete duplication methods provide limited control over the model’s exposure to different classes and do not encode clinical importance.
While data resampling remains the dominant strategy for mitigating imbalance~\cite{oversampling_most_dominant_imbalance_mitigation_and_imbalance_in_cnns_oversampling,resampling_common_and_weighting_difficulties_under_imbalance_noisy_and_loss_weights_cause_instability}, alternative approaches modify the loss via class-weighting schemes~\cite{AsymmetricLossMultilabel}.
Under extreme imbalance, however, such methods may yield unbounded gradients~\cite{unstable_optimization_loss_functions} or amplify noisy gradients from rare classes~\cite{resampling_common_and_weighting_difficulties_under_imbalance_noisy_and_loss_weights_cause_instability}, resulting in unstable optimization~\cite{resampling_common_and_weighting_difficulties_under_imbalance_noisy_and_loss_weights_cause_instability,unstable_optimization_loss_functions}. Conversely, data-level solutions address imbalance more stably by preserving the optimization landscape and gradient dynamics~\cite{resampling_common_and_weighting_difficulties_under_imbalance_noisy_and_loss_weights_cause_instability,unstable_optimization_loss_functions}.
We introduce the Multilabel Class-Prioritized Probabilistic Sampler (ML-CLAPPS), which formulates multilabel sampling as a continuous optimization problem. By learning a distribution over training instances that aligns class exposure with a user-defined importance vector (\iv), it provides a principled approach for addressing both class imbalance and clinical prioritization in multilabel ECG classification.

\section{Methods}

\subsection{Defining the ML-CLAPPS Optimization Problem}

Existing approaches for mitigating multilabel imbalance typically rely on discrete resampling strategies~\cite{LP_ROS,CompareAll,more_duplication_strategies,imbalanced_boosting_and_data_gen,oversampling_most_dominant_imbalance_mitigation_and_imbalance_in_cnns_oversampling,Ando2017DeepOF,undersampling}, such as duplicating instances associated with rare classes.
In contrast, we cast multilabel sampling as a continuous optimization problem, where each training instance is assigned a sampling probability instead of being duplicated a discrete number of times. This probabilistic formulation enables fine-grained control over model exposure to different classes.

\subsubsection{Notation and Expected Class Exposure}

Let $\mathbf{Y} \in \{0,1\}^{m \times n} = [\mathbf{y}_1, \dots, \mathbf{y}_n]$ be the label matrix for $n$ training instances, where each column $\mathbf{y}_j$ is a binary indicator vector encoding the presence of the $m$ possible classes in the $j$-th training instance, such that $Y_{ij} = 1$ if class $i$ is present and $0$ otherwise.
We seek a sampling distribution $\mathbf{p} = [p_1, \dots, p_n]^\top$, where $p_j$ is the probability of drawing the $j$-th instance. Let $\mathbf{y}$ be a random vector representing the labels of an instance sampled according to $\mathbf{p}$. The expected class exposure $\mathbf{d}$ is defined as the expectation of this random vector:
$
    \mathbf{d} = \mathbb{E}[\mathbf{y}] = \sum_{j=1}^{n} p_j \mathbf{y}_j = \mathbf{Y}\mathbf{p} \in [0,1]^m,
$
where $d_i$ denotes the probability that class $i$ appears in an instance sampled according to $\mathbf{p}$. 
Although each $d_i \in [0,1]$, $\mathbf{d}$ is not a probability distribution.

\subsubsection{Exposure-Importance Alignment Term}

We aim to explicitly control class exposure ratios during training by introducing an importance vector (\iv) $\mathbf{w} \in \mathbb{R}^m_{>0}$ that encodes user-defined class priorities. We say that an expected class exposure vector $\mathbf{d}$ is perfectly aligned with an \iv\ $\mathbf{w}$ if they are positively collinear, i.e., if there exists $c > 0$ such that $\mathbf{d} = c\mathbf{w}$. This implies that the relative exposure of any two classes $i$ and $j$ exactly matches their relative importance:
    $
        \frac{d_i}{d_j} = \frac{w_i}{w_j}, \quad \forall i, j \in \{1, \dots, m\}.
    $
Intuitively, $\mathbf{w}$ defines the target exposure ratios: assigning class $i$ twice the importance of class $j$ ($w_i / w_j = 2$) requires its expected training exposure to be twice as high ($d_i / d_j = 2$).
This ensures that the training distribution explicitly reflects the user-defined priorities.
In practice, perfect alignment may be infeasible due to structural constraints from label co-occurrence. For example, if a rare condition always co-occurs with a more frequent one, assigning the rare condition a higher importance cannot be realized.
To penalize deviations from perfect alignment, we first normalize the \iv\ as $\hat{\mathbf{w}} = \mathbf{w} / |\mathbf{w}|_1$.
Given the expected class exposure $\mathbf{d} = \mathbf{Y}\mathbf{p}$, we construct a target exposure vector by computing the total exposure induced by $\mathbf{p}$, $\mathbf{1}^T \mathbf{d} = \mathbf{1}^T \mathbf{Y}\mathbf{p}$, and redistributing this total according to the proportions specified by $\hat{\mathbf{w}}$. This yields the target exposure vector $(\mathbf{1}^T \mathbf{Y}\mathbf{p})\,\hat{\mathbf{w}}$, which preserves the desired relative proportions while matching the overall exposure scale induced by $\mathbf{p}$. The exposure-importance alignment loss term is thus formulated as $\frac{1}{m} \|\mathbf{Y}\mathbf{p} - (\mathbf{1}^T \mathbf{Y}\mathbf{p})\hat{\mathbf{w}}\|_2^2$. To simplify this term, note that since $\mathbf{1}^T \mathbf{Y}\mathbf{p}$ is a scalar, it is commutative with the vector $\hat{\mathbf{w}}$, so $(\mathbf{1}^T \mathbf{Y}\mathbf{p})\hat{\mathbf{w}} = \hat{\mathbf{w}}(\mathbf{1}^T \mathbf{Y}\mathbf{p})$. Furthermore, because matrix multiplication is associative, we have $\hat{\mathbf{w}}(\mathbf{1}^T \mathbf{Y}\mathbf{p}) = (\hat{\mathbf{w}}\mathbf{1}^T \mathbf{Y})\mathbf{p}$. Factoring out $\mathbf{p}$ and defining $\mathbf{M} = \mathbf{Y} - \hat{\mathbf{w}}\mathbf{1}^T \mathbf{Y}$, we obtain the final form of the loss: $\frac{1}{m} \|(\mathbf{Y} - \hat{\mathbf{w}}\mathbf{1}^T \mathbf{Y})\mathbf{p}\|_2^2 = \frac{1}{m} \|\mathbf{M}\mathbf{p}\|_2^2 = \frac{1}{m} \mathbf{p}^T \mathbf{M}^T \mathbf{M} \mathbf{p}$.
This term encourages the learned sampling distribution to achieve exposure-importance alignment by matching expected class exposure to predefined importance proportions.

\subsubsection{Dataset Utilization Regularization}

To prevent degenerate solutions in which the sampling distribution collapses onto a small subset of instances, we introduce a dataset utilization regularization term:
$
\frac{1}{n}\left\|\mathbf{p} - \frac{1}{n}\mathbf{1}\right\|_2^2
=
\frac{1}{n}\left( \mathbf{p}^T\mathbf{p}
- \frac{2}{n}\mathbf{1}^T\mathbf{p}
+ \frac{1}{n^2}\mathbf{1}^T\mathbf{1} \right).
$
This term encourages the sampling distribution to remain close to uniform, promoting broad utilization of the dataset. It acts as a stabilizing regularizer that counterbalances the exposure objective: without it, the exposure term can be easily minimized by concentrating probability mass on a small subset of instances to approach perfect alignment, while neglecting the rest of the dataset.

\subsubsection{Final ML-CLAPPS Objective}
We define $\gamma > 0$ to balance the exposure-importance alignment term against the dataset utilization term. By expanding the squared norms, the objective reformulates into a quadratic program (QP):

\begin{equation}
\min_{\mathbf{p} \in \Delta^{n-1}} \mathcal{L}(\mathbf{p}) = 
\underbrace{\frac{1}{m}\|\mathbf{M}\mathbf{p}\|_2^2}_{\text{Alignment}} + 
\underbrace{\frac{\gamma}{n} \left\|\mathbf{p}-\frac{1}{n}\mathbf{1}\right\|_2^2}_{\text{Dataset Utilization}} = 
\frac{1}{2}\mathbf{p}^\top\mathbf{H}\mathbf{p} + \left( -\frac{2\gamma}{n^2} \mathbf{1} \right)^\top \mathbf{p} + \frac{\gamma}{n^2}
\label{eq:mlclapps_qp}
\end{equation}

where $\Delta^{n-1} = \{ \mathbf{p} \in \mathbb{R}^n \mid \mathbf{1}^T \mathbf{p} = 1, \mathbf{p} \succeq 0 \}$ is the probability simplex, a compact convex set~\cite{rudin1976principles,boyd2004convex}. The Hessian $\mathbf{H} = \frac{2}{m} \mathbf{M}^T \mathbf{M} + \frac{2\gamma}{n} \mathbf{I}$ is the sum of a symmetric positive semidefinite matrix $\frac{2}{m} \mathbf{M}^T \mathbf{M}$ and a symmetric positive definite matrix $\frac{2\gamma}{n} \mathbf{I}$, and is therefore strictly positive definite~\cite{boyd2004convex}.
Hence, Eq.~\eqref{eq:mlclapps_qp} is a strictly convex QP over a compact set, which ensures the existence of a unique global minimizer $\mathbf{p}^\star$~\cite{rudin1976principles,boyd2004convex}.
Moreover, since $\mathbf{1}^\top\mathbf{p}=1$ for all $\mathbf{p}\in\Delta^{n-1}$, the linear term $-\frac{2\gamma}{n^2}\mathbf{1}^\top\mathbf{p} = -\frac{2\gamma}{n^2}$ is constant over the feasible set. Removing constant terms does not affect the minimizer~\cite{boyd2004convex}, yielding the purely quadratic form $\mathbf{p}^\star=\arg\min_{\mathbf{p}\in\Delta^{n-1}}\tfrac{1}{2}\mathbf{p}^\top\mathbf{H}\mathbf{p}$.
The problem is solved once before training, after which $\mathbf{p}^\star$ is fixed and used to sample training instances at each epoch.

\subsection{Efficient Solution of the ML-CLAPPS Problem}

Boyd and Vandenberghe~\cite{boyd2004convex} (Section 4.4.1) describe convex quadratic programs with linear inequality constraints, such as Eq.~\eqref{eq:mlclapps_qp}, as constrained regression problems for which no simple analytical solution exists~\cite{boyd2004convex}. The inequality constraints force a combinatorial search for active sets, necessitating iterative methods~\cite{boyd2004convex}.
Therefore, we solve the QP using a primal-dual interior-point method, which has polynomial-time complexity in problem dimensions and logarithmic dependence on the desired precision~\cite{Nesterov1994InteriorpointPoly,boyd2004convex}, implemented using the Clarabel solver~\cite{goulart2024clarabelinteriorpointsolverconic}.

\begin{figure}[t]
    \centering
    \includegraphics[width=0.8\linewidth]{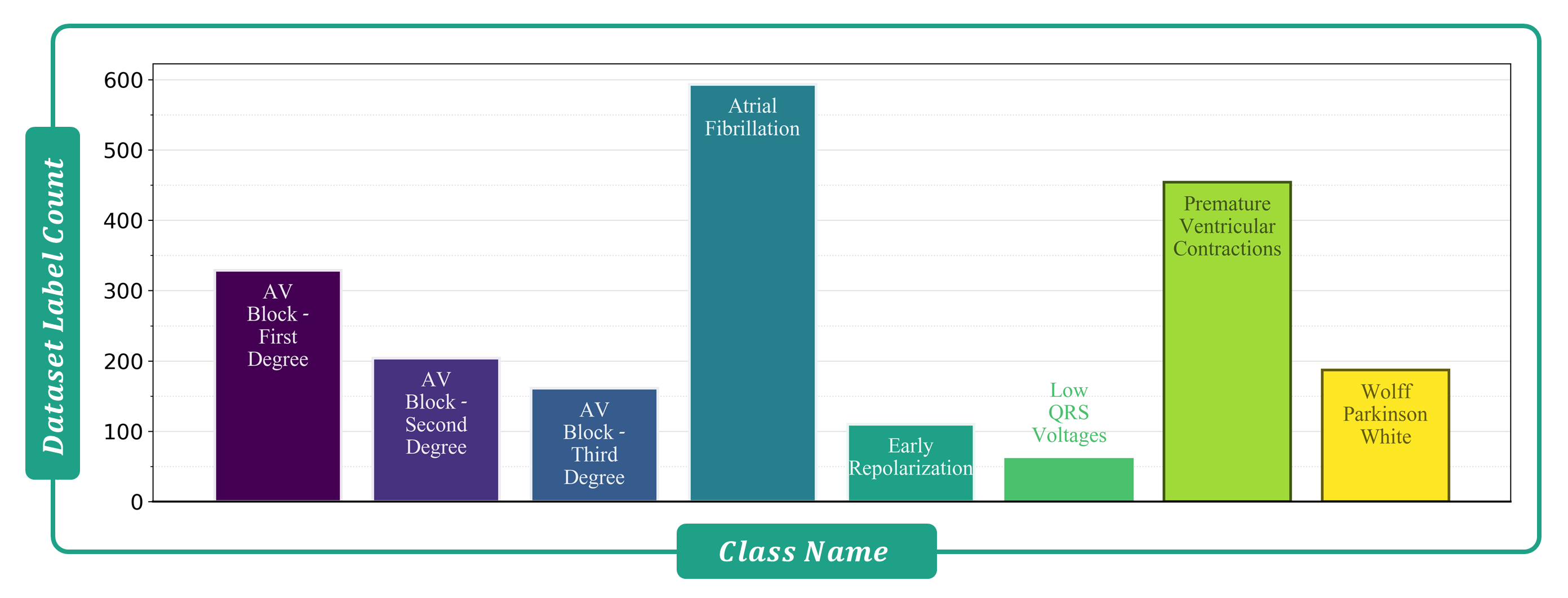}
    \caption{Class distribution of the Mobile dataset, illustrating dataset imbalance.}

    \label{fig:multilabel:labels-distribution}
\end{figure}

\section{Experiments}
\subsection{Experimental Setup}
\label{sec:experimental-setup}

To demonstrate clinical applicability, we collected the Multilabel Mobile dataset, comprising 5,197 multilabel ECG photographs captured via mobile cameras across five hospitals.
The dataset was split into 85\% training and 15\% test sets.
The label distribution in Fig.~\ref{fig:multilabel:labels-distribution} highlights the severe multilabel imbalance present in the dataset.
All experiments utilize a ResNet18 visual backbone~\cite{resnet} for multilabel image classification. To ensure a fair comparison, we fixed the architecture, optimization settings, and the number of training epochs, varying only the sampling strategy.
We report F1 score, Matthews correlation coefficient (MCC)~\cite{MCC}, accuracy, and computational time overhead, categorized into one-time initialization time and recurring per-epoch sampling latency.
Under severe imbalance, accuracy can be misleading due to the dominance of negative samples. For example, if a condition appears in only 4\% of the data, a model predicting its absence 100\% of the time achieves 96\% accuracy while failing to detect any positive cases, a phenomenon known as the accuracy paradox~\cite{accuracy_paradox}. Consequently, we prioritize F1 and MCC~\cite{MCC,eval_metrics}, as they remain informative under imbalance: F1 is sensitive to missed positives, whereas MCC incorporates all four confusion-matrix terms and yields a score of zero for constant classifiers when both positive and negative instances are present.
We compare against LP~ROS~\cite{LP_ROS,CompareAll}, ClusterClip~\cite{ClusterClip} (24 labelset clusters), LeastSampledClassSampler~\cite{issamemari2021sampler}, and UniformSampler, a naive baseline that samples instances from a uniform distribution, ignoring imbalance.

\begin{figure}
    \centering
    \includegraphics[width=\linewidth]{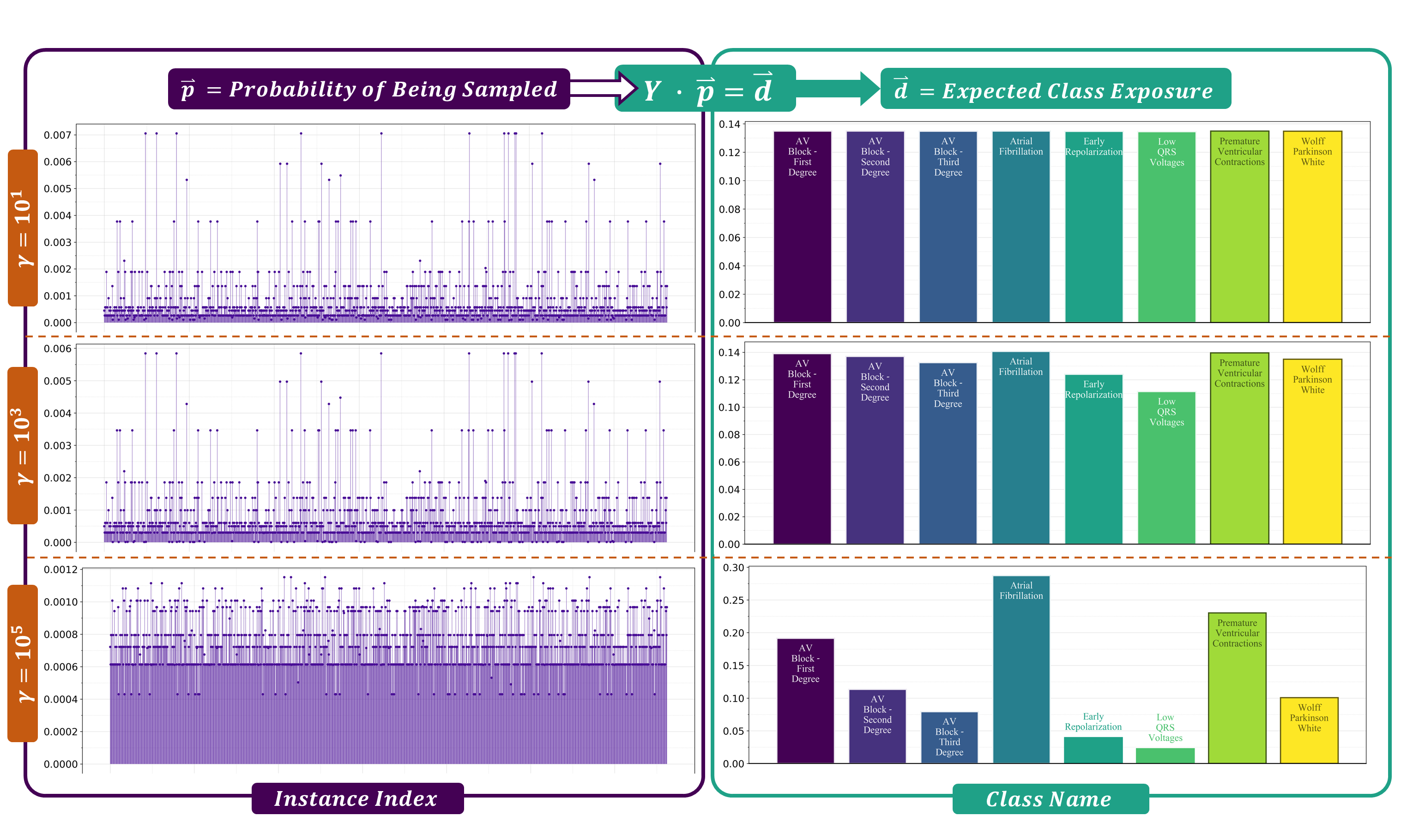}
\caption{Optimal $\mathbf{p}^\star$ (left) and $\mathbf{d}$ (right) from ML-CLAPPS under varying $\gamma$ with uniform \iv. Larger $\gamma$ improves dataset utilization via more uniform sampling, at the cost of shifting exposure toward the original imbalanced label distribution (Fig.~\ref{fig:multilabel:labels-distribution}).}
    \label{fig:multilabel:pd-vs-gamma}
\end{figure}

\subsection{Analysis of ML-CLAPPS Behavior}

\subsubsection{Effect of the Trade-off Parameter}

\begin{figure}[!htbp]
    \centering
    \includegraphics[width=\linewidth]{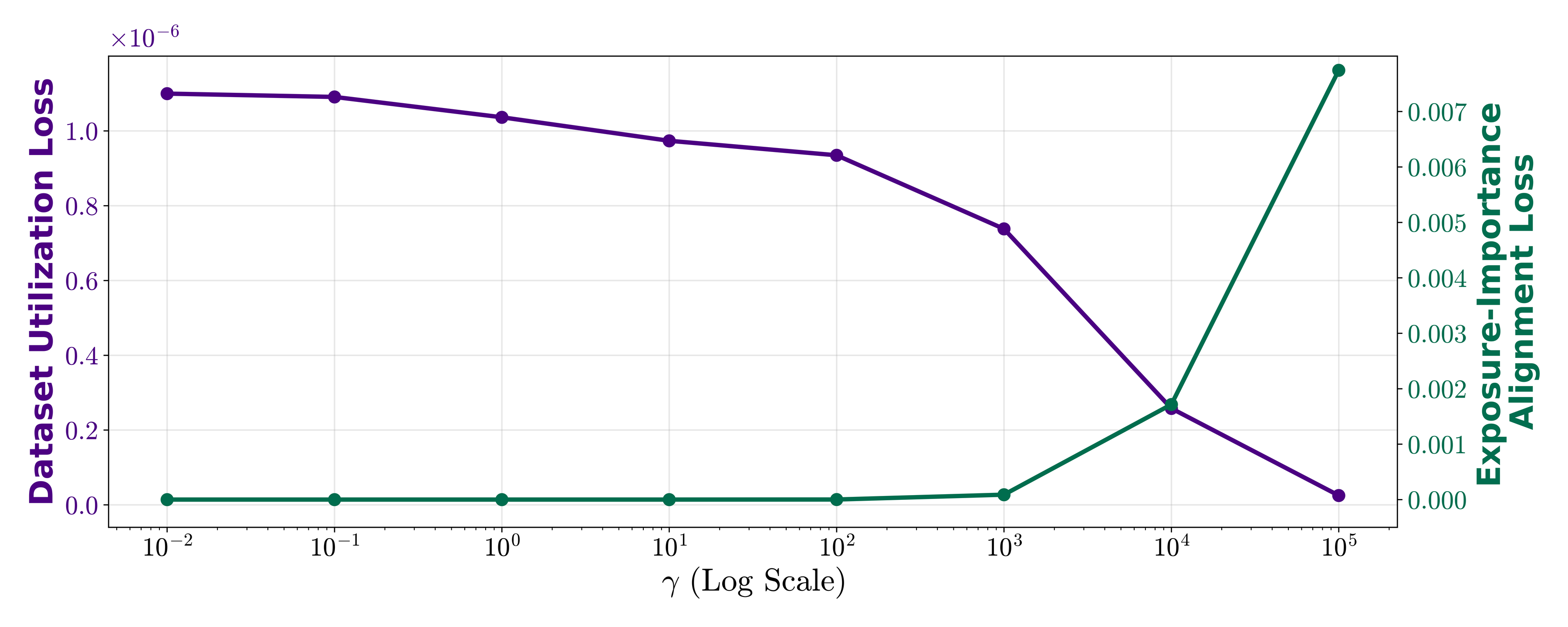}
\caption{Effect of $\gamma$ on dataset utilization loss (left axis) and alignment loss (right axis), derived from solving ML-CLAPPS for each $\gamma$ under a uniform \iv. Lower values denote improved utilization and closer exposure-importance alignment.}
    \label{fig:multilabel:gamma-vs-loss-terms}
\end{figure}

For a uniform \iv, Fig.~\ref{fig:multilabel:pd-vs-gamma} illustrates the effect of the trade-off parameter $\gamma$ on $\mathbf{p}^\star$ and $\mathbf{d}$, which is reflected in the corresponding dataset utilization and alignment losses (Fig.~\ref{fig:multilabel:gamma-vs-loss-terms}).
For $\gamma=10^1$, $\mathbf{d}$ aligns with the uniform \iv, but $\mathbf{p}^\star$ is highly non-uniform, indicating poor dataset utilization.
For $\gamma=10^5$, $\mathbf{p}^\star$ becomes nearly uniform, while $\mathbf{d}$ drifts toward the original label distribution (Fig.~\ref{fig:multilabel:labels-distribution}), reintroducing imbalance.
We use $\gamma=10^3$ in all experiments, as it achieves reasonably uniform sampling while retaining close alignment of $\mathbf{d}$ with the desired \iv\ (Fig.~\ref{fig:multilabel:pd-vs-gamma}), thereby reducing the utilization regularization loss while keeping the alignment loss extremely low (Fig.~\ref{fig:multilabel:gamma-vs-loss-terms}).

\subsubsection{Exposure Alignment with User-Defined Importance}

\begin{figure}
    \centering
    \includegraphics[width=0.8\linewidth]{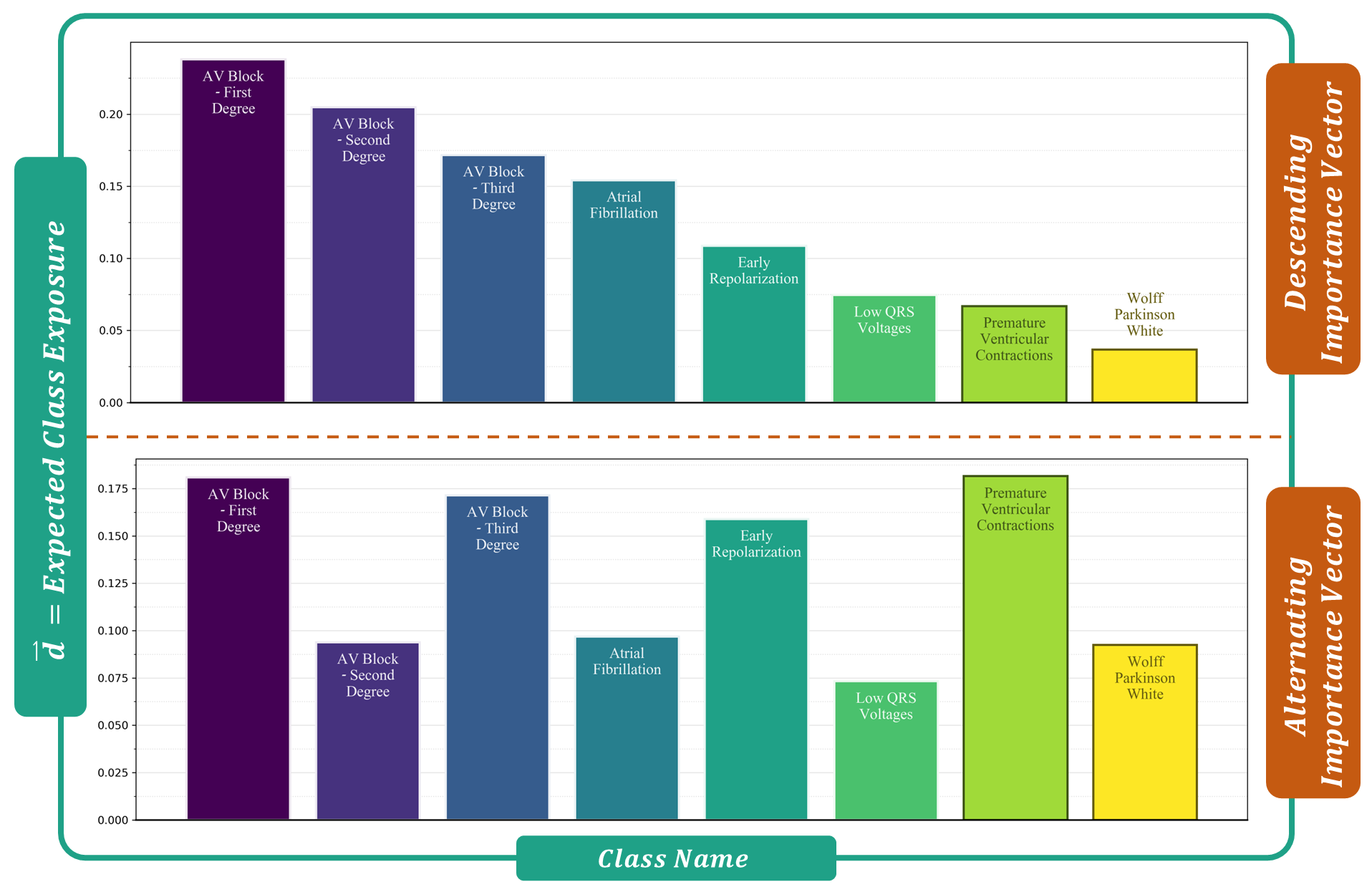}
\caption{Expected class exposure $\mathbf{d}$ obtained by solving ML-CLAPPS for different \ivs\ using $\gamma=10^3$. Top: descending importance. Bottom: alternating importance. Expected exposure closely follows the prescribed class priorities, mirroring the user-defined \ivs.}
    \label{fig:importance_vectors_demo}
\end{figure}

We evaluated ML-CLAPPS’s ability to align class exposure with user-defined priorities by solving the objective for illustrative descending 
($\mathbf{w} = [8, 7, 6, 5, 4, 3, 2, 1]$) and alternating ($\mathbf{w} = [2, 1, 2, 1, 2, 1, 2, 1]$)
\ivs. As shown in Fig.~\ref{fig:importance_vectors_demo}, the resulting expected class exposure $\mathbf{d}$ closely mirrors the prescribed importance ratios, confirming that the framework effectively translates user priorities into targeted training emphasis on selected conditions, even under complex label co-occurrences.

\begin{table}[!t]
    \centering
        \caption{Comparative performance of sampling strategies on the Mobile ECG dataset. ML-CLAPPS outperforms prior methods. Table reports mean $\pm$ SD across four runs.}
        \label{tab:multilabel-results}
    \setlength{\tabcolsep}{1.5pt}%
    \fontsize{8}{9.5}\selectfont
    \begin{tabular*}{\textwidth}{@{\extracolsep{\fill}}l|ccc||cc}
        \toprule
        \textbf{Method} 
        & \multicolumn{3}{c}{\textbf{Classification Performance}} 
        & \multicolumn{2}{c}{\textbf{Overhead (sec)} $\downarrow$} \\
        \cmidrule(lr){2-4} \cmidrule(lr){5-6}
        & \textbf{F1} $\uparrow$ 
        & \textbf{MCC} $\uparrow$ 
        & \textbf{Accuracy} $\uparrow$
        & \textbf{Init} 
        & \textbf{Per-Epoch} \\
        \midrule
        UniformSampler (Baseline) 
        & 0.281$\pm$.02 
        & 0.301$\pm$.02 
        & 0.954$\pm$.001
        & \textbf{0} 
        & \textbf{0} \\
        LP ROS~\cite{LP_ROS,CompareAll} 
        & 0.291$\pm$.01 
        & 0.315$\pm$.01 
        & 0.949$\pm$.001
        & 0.05$\pm$.01
        & \textbf{0} \\
        ClusterClip Sampling~\cite{ClusterClip} 
        & 0.299$\pm$.02 
        & 0.344$\pm$.03 
        & 0.953$\pm$.002
        & 0.01$\pm$.00
        & 0.01$\pm$.00 \\
        LeastSampledClass~\cite{issamemari2021sampler} 
        & 0.325$\pm$.02 
        & 0.365$\pm$.02 
        & 0.954$\pm$.001
        & \textbf{0} 
        & 0.49$\pm$.05 \\
        \midrule
        ML-CLAPPS (\iv=\textsc{LinR2F}) 
        & 0.347$\pm$.01 
        & 0.363$\pm$.01 
        & 0.954$\pm$.001
        & 1.17$\pm$.20
        & \textbf{0} \\
        ML-CLAPPS (\iv=uniform) 
        & \textbf{0.386}$\pm$.01 
        & \textbf{0.421}$\pm$.01 
        & \textbf{0.956}$\pm$.001
        & 1.06$\pm$.09
        & \textbf{0} \\
        \bottomrule
    \end{tabular*}
\end{table}

\subsection{Performance under Class Imbalance and Prioritization}

\begin{figure}
    \centering
    \includegraphics[width=0.9\linewidth]{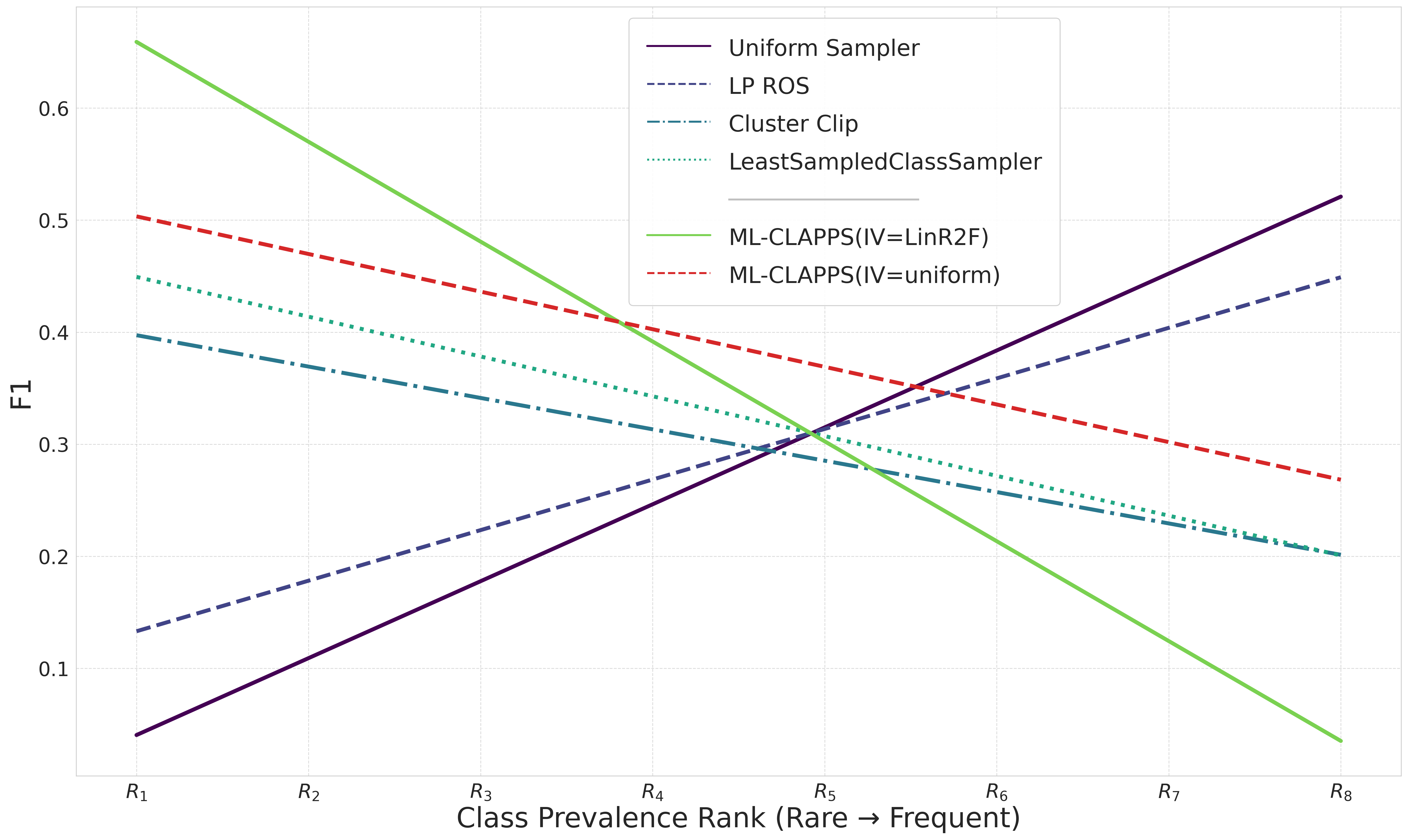}
\caption{The x-axis shows class prevalence from rare (left) to frequent (right), where $R_k$ indicates the kth most rare class.
The y-axis reports mean F1 across four runs.
For each sampling method, a fitted linear regression line is shown to highlight performance trends across class frequencies. \textsc{LinR2F} \iv\ demonstrates targeted condition prioritization.}
    \label{fig:multilabel:f1-vs-prevalence}
\end{figure}

ML-CLAPPS was evaluated using two \ivs: a uniform \iv\ to address general imbalance, and a Linear Rare-to-Frequent (\textsc{LinR2F}) \iv. The \textsc{LinR2F} vector assigns linearly decaying importance weights, ranging from 4.5 for the rarest condition (Low QRS Voltages) to 1.0 for the most frequent (Atrial Fibrillation), and is designed to examine how prioritized importance correlates with performance gains on targeted conditions.
We assess imbalance mitigation using macro-averaged performance across conditions (Table~\ref{tab:multilabel-results}).
Accuracy differs only marginally between methods, as it is less informative under severe imbalance (Section~\ref{sec:experimental-setup}) due to the accuracy paradox~\cite{accuracy_paradox}.
ML-CLAPPS with a uniform \iv\ achieves superior performance across all classification metrics, substantially improving the imbalance-robust F1 and MCC scores, demonstrating its ability to mitigate imbalance effectively.
Although all methods incur only seconds of overhead on this dataset (Table~\ref{tab:multilabel-results}), in large-scale datasets and long training schedules with many epochs, where overheads may become non-negligible, ML-CLAPPS is expected to be more scalable for two reasons.
First, it uniquely combines high classification performance with zero per-epoch overhead (Table~\ref{tab:multilabel-results}) by pre-calculating the optimal distribution $\mathbf{p}^\star$ once in polynomial time~\cite{Nesterov1994InteriorpointPoly,boyd2004convex} and using it to sample training instances across all epochs. This avoids the recurring latency seen in methods like LeastSampledClassSampler~\cite{issamemari2021sampler} and ClusterClip~\cite{ClusterClip}, where overhead scales with the number of epochs.
Second, the iterative nature of the interior-point method supports early termination when a near-optimal solution suffices, providing a flexible trade-off between sampler initialization overhead and solution precision that other samplers lack.
Figure~\ref{fig:multilabel:f1-vs-prevalence} illustrates performance trends across condition prevalences.
ML-CLAPPS with a uniform \iv\ achieves consistently high performance across the full prevalence spectrum.
LP ROS~\cite{LP_ROS,CompareAll} and UniformSampler improve only as condition prevalence increases and yield near-zero F1 on the rarest conditions, highlighting their inability to address imbalance driven by frequent labels.
The \textsc{LinR2F} \iv\ further demonstrates the framework’s flexibility, as performance across the prevalence spectrum follows the assigned importance weights, achieving the highest performance on the rarest conditions and gradually decreasing as conditions become more frequent, as \textsc{LinR2F} prescribes.
This highlights ML-CLAPPS's ability to enable clinical prioritization of targeted conditions through a user-defined \iv.

\section{Conclusion}

We introduced ML-CLAPPS, a scalable probabilistic framework for imbalanced multilabel learning with explicit class prioritization.
We formulated multilabel sampling as a convex QP that balances exposure-importance alignment with dataset utilization, solving it once in polynomial time with zero per-epoch overhead.
On a real-world imbalanced ECG image dataset, ML-CLAPPS outperforms existing samplers in macro-averaged metrics and across most of the prevalence spectrum, effectively mitigating imbalance.
We showed that user-defined \ivs\ directly shape class exposure and translate into controlled performance trade-offs across classes, underscoring the applicability of ML-CLAPPS to clinical settings where prioritization of significant conditions is essential.

\bibliographystyle{splncs04}
\bibliography{references}

\end{document}